\documentclass[]{article} 
\usepackage[]{iclr2025_conference,times}

\usepackage{amsmath,amsfonts,bm}

\def\eqref#1{equation~\ref{#1}}

\def\1{\bm{1}}

\DeclareMathAlphabet{\mathsfit}{\encodingdefault}{\sfdefault}{m}{sl}
\SetMathAlphabet{\mathsfit}{bold}{\encodingdefault}{\sfdefault}{bx}{n}

\usepackage[hidelinks]{hyperref}
\usepackage{url}
\usepackage{soul,color}

\usepackage{multirow}
\usepackage{booktabs}
\usepackage{graphicx}
\usepackage{float}
\usepackage{algorithm}
\usepackage{algpseudocode}
\usepackage{caption}
\usepackage{subcaption}
\usepackage{amssymb}
\usepackage{soul}

\usepackage[utf8]{inputenc} 
\usepackage[T1]{fontenc}    
\usepackage{hyperref}       
\usepackage[capitalize]{cleveref}       
\usepackage{url}            
\usepackage{wrapfig,booktabs}       
\usepackage{amsfonts}       
\usepackage{nicefrac}       
\usepackage{microtype}      
\usepackage{xcolor}
\usepackage{amsmath}
\usepackage{mathtools}
\usepackage{pgf-pie}
\usepackage{graphicx}
\usepackage{pgfplots}
\usetikzlibrary{patterns}
\pgfplotsset{compat=1.18}
\usepackage{graphicx}
\usepackage{pythonhighlight}
\usepackage{bbding}
\usepackage{listings}
\usepackage{geometry}
\usepackage{array}
\usepackage{fvextra}
\usepackage{listings}
\usepackage{colortbl}

\usepackage{tikz}
\usepackage{fontawesome5}

\let\realcite\citep
\renewcommand{\cite}[1]{\ifx.#1.\hl{[?]}\else\realcite{#1}\fi}

\crefformat{section}{\S#2#1#3} 
\crefformat{subsection}{\S#2#1#3}
\crefformat{subsubsection}{\S#2#1#3}

\newcommand{\model}{\textsc{NORA}}
\newcommand{\modellong}{\textsc{NORA-Long}}

\title{\model{}: A Small Open-Sourced Generalist Vision Language Action Model for Embodied Tasks}

\author{%
    Chia-Yu Hung\textsuperscript{*}
    , Qi Sun\textsuperscript{*}, Pengfei Hong\textsuperscript{*}\vphantom{\thanks{Equal Contributions.}}\\
    Singapore University of Technology and Design \\
    \And
    \textbf{Amir Zadeh, Chuan Li} \\
    Lambda Labs \\
    \And
    \textbf{U-Xuan Tan, Navonil Majumder, Soujanya Poria} \\
    Singapore University of Technology and Design \\
}

\iclrfinalcopy
\begin{document}

\maketitle

\begin{figure}[ht]
    \centering
    \includegraphics[width=0.6\textwidth]{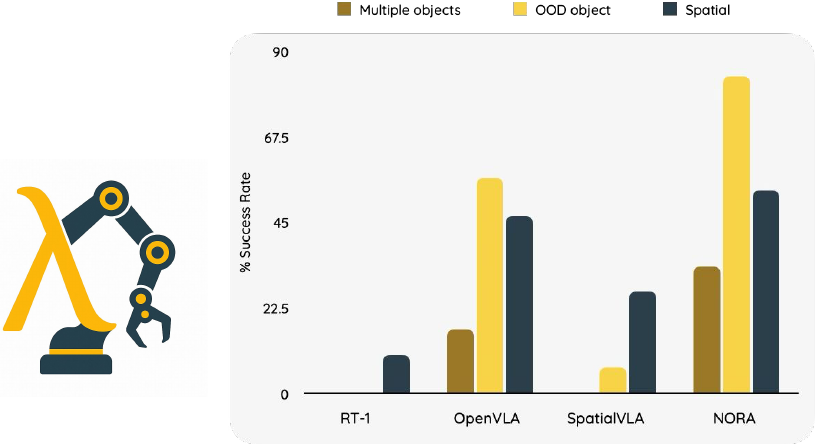}
\end{figure}

\centerline{\textbf{\textcolor{blue}{\texttt{Codes \& Checkpoints:} \url{https://declare-lab.github.io/nora}}}}

\begin{tikzpicture}[remember picture,overlay,shift={(current page.north west)}]
\node[anchor=north west,xshift=0.5cm,yshift=-2.4cm]{\scalebox{1}[1]{\includegraphics[width=2.5cm]{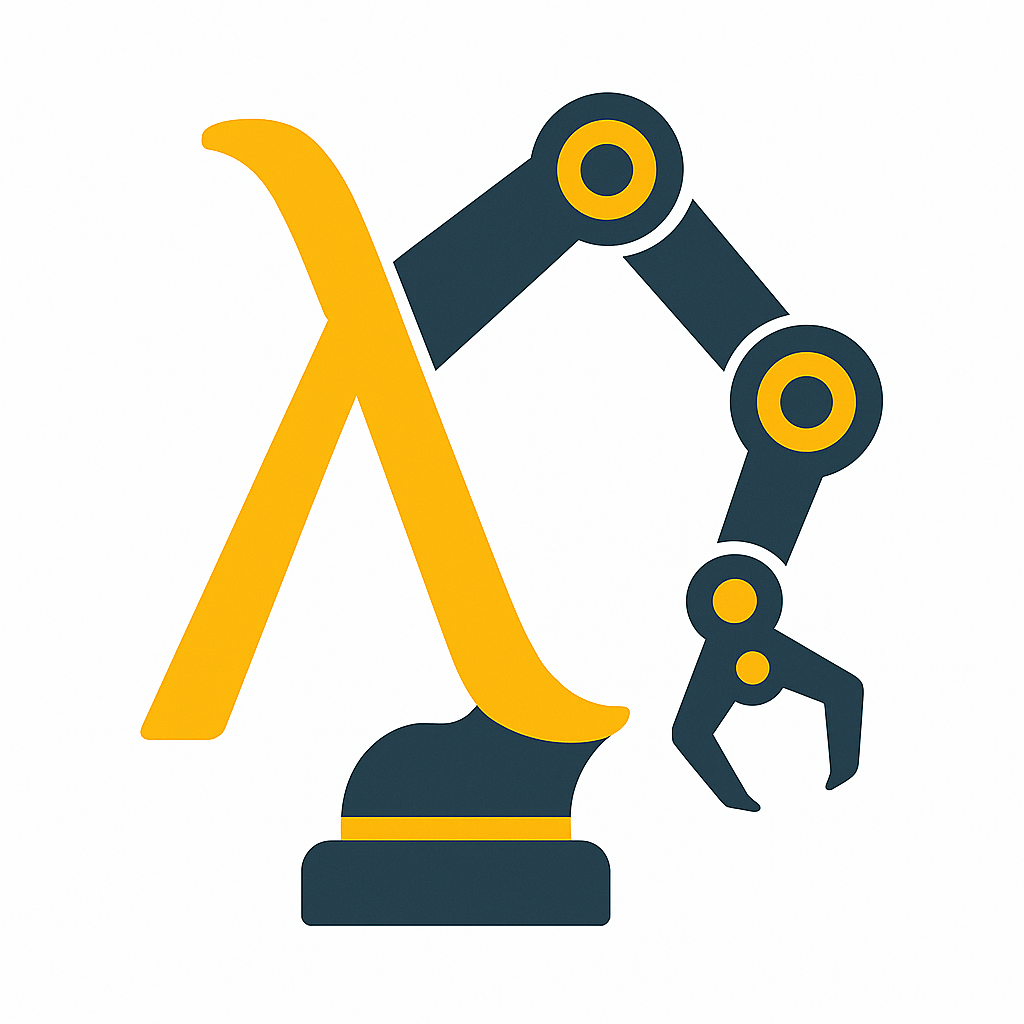}}};
\end{tikzpicture}

\begin{abstract}
Existing Visual-Language-Action (VLA) models have shown promising performance in zero-shot scenarios, demonstrating impressive task execution and reasoning capabilities. However, a significant challenge arises from the limitations of visual encoding, which can result in failures during tasks such as object grasping. Moreover, these models typically suffer from high computational overhead due to their large sizes, often exceeding 7B parameters. While these models excel in reasoning and task planning, the substantial computational overhead they incur makes them impractical for real-time robotic environments, where speed and efficiency are paramount. Given the common practice of fine-tuning VLA models for specific tasks, there is a clear need for a smaller, more efficient model that can be fine-tuned on consumer-grade GPUs.  To address the limitations of existing VLA models, we propose NORA, a 3B-parameter model designed to reduce computational overhead while maintaining strong task performance. NORA adopts the Qwen-2.5-VL-3B multimodal model as its backbone, leveraging its superior visual-semantic understanding to enhance visual reasoning and action grounding. Additionally, our \model{} is trained on 970k real-world robot demonstrations and equipped with the FAST+ tokenizer for efficient action sequence generation. Experimental results demonstrate that NORA outperforms existing large-scale VLA models, achieving better task performance with significantly reduced computational overhead, making it a more practical solution for real-time robotic autonomy.

\end{abstract}

\section{Introduction}

The robotic policy model aims to generate sequences of low-level action policies for robotic systems. Traditional reinforcement learning-based approaches to robotic control often focus on narrowly defined tasks within fixed environments \citep{ma2024survey}. These methods, while effective within their domain, are limited in their ability to generalize beyond specific training tasks, constraining their broader applicability \citep{brohan2023rt1roboticstransformerrealworld}.

In recent years, foundation models for vision and language have emerged as powerful tools, demonstrating exceptional capabilities in scene understanding and task planning \citep{radford2021learning,zhai2023sigmoid,touvron2023llama}. Vision-Language Models (VLMs), in particular, excel at breaking down complex tasks into smaller, manageable steps through chain-of-thought reasoning. These models have shown significant potential in enhancing task planning by leveraging multimodal inputs. However, VLMs are not inherently designed to directly generate policies suitable for specific robotic embodiments, which poses a challenge when applying them to real-world robotic tasks. To bridge this gap, Visual-Language-Action (VLA) models have been developed, utilizing multimodal inputs to generate adaptive and generalized robotic actions for complex, multi-task scenarios \citep{brohan2023rt2visionlanguageactionmodelstransfer,kim2024openvla,octo_2023}.

Despite their success, existing VLA models are typically large-scale, with model sizes approaching 7B parameters, such as OpenVLA \citep{kim2024openvla}, and even larger in methods like TraceVLA \citep{tracevla2024}, ECOT \citep{zawalski2024robotic}, and EMMA-X \citep{sun2024emmaxembodiedmultimodalaction}. These models enhance the reasoning capabilities of robotic systems by incorporating Chain-of-Thought (CoT) mechanisms that combine visual and language understanding to improve task execution accuracy. However, this enhancement comes at the cost of significantly increased computational overhead, as CoT methods require processing intermediate reasoning steps during task execution.

To address these challenges, we introduce the \textbf{N}eural \textbf{O}rchestrator for \textbf{R}obotic \textbf{A}utonomy, \model{}, a 3B-parameter VLA model trained on the Open X-Embodiment dataset \citep{open_x_embodiment_rt_x_2023}. Our goal with \model{} is to reduce computational overhead while maintaining strong task execution capabilities. By leveraging the state-of-the-art open-source multimodal model Qwen-2.5-VL-3B \citep{Qwen2.5-VL}, \model{} achieves a balance between performance and efficiency, enabling more scalable and practical deployment in robotic systems. Furthermore, we employ the FAST+ tokenizer \citep{pertsch2025fastefficientactiontokenization} to discretize continuous action tokens, optimizing action sequence generation for a wide range of robotic tasks. We also demonstrate that with this simple design, we can outperform SpatialVLA in real-world settings, without the need for action grids or spatial embeddings.

Through \model{}, we aim to advance the development of VLA models by offering a scalable solution that combines strong reasoning abilities with efficient execution. We extensively evaluate \model{} across various real-world tasks and the LIBERO simulation benchmark \citep{liu2023libero}. Experimental results show that \model{} achieves significant performance improvements over existing competitive baselines.

Our contribution can be summarized as follows:
\begin{itemize}
    \item We propose \model{}, a 3B-parameter VLA model built upon the Qwen-2.5-VL-3B backbone, incorporating an efficient action decoding strategy that compresses highly correlated action tokens while ensuring robust performance across a variety of robotic tasks.
    
    \item We conduct comprehensive experiments to analyze the impact of different action prediction strategies, including a detailed comparison between single-step and chunked action prediction, demonstrating the effectiveness of our design in improving action generation efficiency.
    \item We open-source the full \model{} framework, including model checkpoints, training strategy, and evaluation protocols, to facilitate reproducibility and promote further research in scalable visual-language-action models for robotics.
\end{itemize}

\section{Preliminaries}
\label{sec:prelims}

\subsection{Vision-Language Models (VLMs)}
Vision-Language Models (VLMs) have become powerful frameworks for image understanding and reasoning, demonstrating the ability to generate text based on visual input, and identifying objects in images. This serves as an excellent choice of backbone for VLAs. VLAs finetuned from pre-trained VLMs significantly benefit from this internet-scale image and text pre-training that these models undergo. This pretraining imparts a rich understanding of visual semantics, enabling VLAs to ground language in the visual world and translate that understanding into meaningful robotic actions. Such grounding facilitates generalization to out-of-distribution instructions and environments. For example,  VLA may generalize from prior visual-language experience to interpret and execute an instruction like ``\textit{pick up a toy}'' in a previously unseen scene, despite not having encountered the exact phrase or context during training VLA training.

Recent Vision-Language Models (VLMs) comprise an image encoder \cite{dinov2, siglip}, a Large Language Model (LLM) backbone \cite{touvron2023llama}, and a projection network that maps visual representations into a shared embedding space. This architecture enables the LLM to effectively reason over both text and image modalities. The pretraining of VLMs typically leverages diverse multi-modal datasets comprising interleaved image-text pairs, visual knowledge sources, object grounding, spatial reasoning, multi-modal question-answering datasets.
Our work builds on the Qwen2.5-VL model \cite{Qwen2.5-VL}, a state-of-the-art open-source VLM. A notable feature of Qwen2.5-VL is its use of native image resolution during training, which aims to enhance the model's perception of real-world scale and spatial relationships. This approach enables more accurate understanding of object sizes and positions, leading to improved performance on tasks such as object detection and localization. We hypothesize that we can leverage the grounding and spatial ability of Qwen 2.5-VL in building VLAs, which can be helpful for robot control.

\subsection{Vision-Language-Action Models (VLAs)}
Despite their strengths, VLMs are not inherently designed to directly generate policies applicable to specific embodiment configurations in robotics. This limitation has spurred the emergence of \textbf{Visual-Language-Action (VLA) models}, which bridge this gap by leveraging multimodal inputs---combining visual observations and language instructions---to produce adaptive and generalized robotic actions across diverse, multi-task scenarios. VLA models can be broadly categorized into two types based on their action modeling methods: \textbf{continuous-action models} \cite{octo_2023}, which typically employ diffusion processes to generate smooth trajectories in continuous action spaces, and \textbf{discrete-token models} \cite{brohan2023rt, brohan2023rt1roboticstransformerrealworld, kim2024openvla, sun2024emmaxembodiedmultimodalaction}, where robotic actions are represented as sequences of discrete tokens. In the discrete token-based VLA formulation for imitation learning, the robot's state at a given time $t$ is characterized by a multimodal observation including visual images $I_t$, textual instructions $L_t$, and prior state context $S_t$. The goal is to predict a sequence of discrete tokens $A_t$, representing actions executable by the robot. Formally, the imitation learning policy model $\pi_{\theta}(A_t \mid I_t, L_t, S_t)$ is trained to replicate expert-provided action sequences, enabling the robot to generalize learned behaviors to novel scenarios guided by visual-language prompts.

\subsection{Action tokenization}
In robotic systems, actions are typically represented as continuous control signals across multiple degrees of freedom (DoFs), such as translation in $(x, y, z)$ and rotation in roll, pitch, and yaw. To enable compatibility with transformer-based language backbones, it is common to discretize these continuous actions via binning approaches \cite{brohan2023rt1roboticstransformerrealworld, brohan2023rt}. This process maps each dimension of a robot action to one of 256 discrete bins using a quantile-based strategy, ensuring robustness against outliers while maintaining sufficient granularity. OpenVLA \cite{kim2024openvla} incorporates these action tokens into the language model's vocabulary by overwriting the 256 least-used tokens in the LLaMA tokenizer, enabling next-token prediction over action sequences. To further improve pretraining efficiency, we adopt a fast tokenization method \cite{pertsch2025fastefficientactiontokenization} which applies discrete cosine transform (DCT) across the action dimensions at each timestep. This decorrelates joint action components and enables the use of byte-pair encoding (BPE) to compress them into shorter, more token-efficient sequences. The resulting representation reduces vocabulary size and accelerates convergence, while aligning the structure of action data with language-model-friendly token statistics. During inference, \model{} uses about 8.3GB of GPU memory.

\section{\model{}}
\label{sec:nora}

We introduce \textbf{N}eural \textbf{O}rchestrator for \textbf{R}obotic \textbf{A}utonomy, \model{}, a 3B-parameter Vision-Language-Action (VLA) model trained on the Open X-Embodiment dataset \cite{open_x_embodiment_rt_x_2023}. Built upon an existing Vision-Language Model (VLM), \model{} leverages its strong general world knowledge, multi-modal reasoning, representation learning, and instruction-following capabilities. Particularly, we adopt the state-of-the-art open-source multi-modal model Qwen-2.5-VL-3B~\cite{Qwen2.5-VL} as the VLM backbone for \model{}, due to its excellent balance between performance and efficiency at this scale. On the other hand, we utilize FAST+ tokenizer \cite{pertsch2025fastefficientactiontokenization} (see \cref{sec:prelims}) to discretize the continuous action tokens, given its proven efficacy across a wide range of action sequences, including single-arm, bi-manual, and mobile robot tasks, making it a strong off-the-shelf choice for training autoregressive VLA models.

\subsection{Architecture}

\begin{figure}
    \centering
    \includegraphics[width=\linewidth]{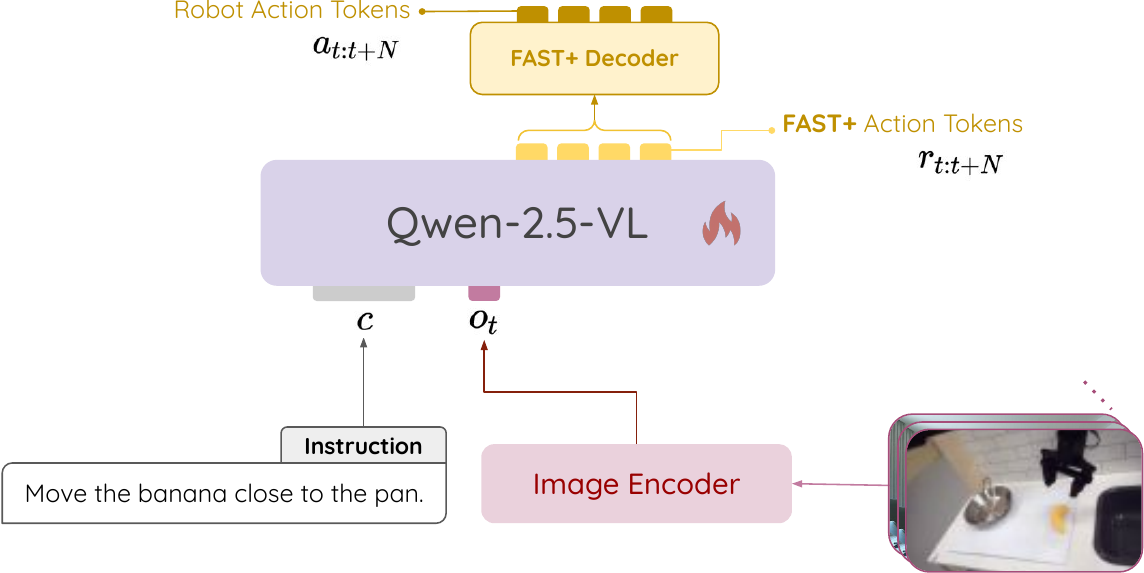}
    \caption{The overall architecture and inference flow of \model{}.}
    \label{fig:nora-arch}
\end{figure}

Our model \model{}, as shown in \cref{fig:nora-arch}, leverages a pre-trained Vision-Language Model (VLM) denoted as $\mathcal{M}$, to auto-regressively predict an action chunk encoding the future actions from time $t$ to $t+N$, denoted as $a_{t:t+N} = [a_t, \ldots, a_{t+N}]$. The input to $\mathcal{M}$ consists of a natural language task instruction $c$ and a visual observation of $n$ frames $o_t = [I_t^1, \ldots, I_t^n]$ at time $t$, which are concatenated to form the overall input $X_t = [o_t, c]$. The action chunk $a_{t:t+N}$ is represented by a sequence of discrete tokens $R = [r_t, \ldots, r_{t+N}]$, encoded using FAST+ robotic tokenizer at training time. The VLM $\mathcal{M}$ predicts this action chunk by autoregressively generating its token sequence $R$ conditioned on $X_t$:
\begin{flalign}
    r_{t:t+N} \sim \mathcal{M}_\theta(r \mid c, o_t), \\
    a_{t:t+N} \leftarrow \text{FAST+}_\text{decode}(r_{t:t+N}).
\end{flalign}



We chose the state-of-the-art open-source VLM Qwen-2.5-VL~\cite{Qwen2.5-VL} as the backbone due to its small 3B parameter size. Additionally, we augmented the vocabulary of VLM tokenizer by 2048 additional tokens introduced by the FAST+ tokenizer. We kept the observations $o_t$ to single visual frame. We chose the action chunk size to be 1. Subsequently, we trained the \model{} using a standard language modeling objective of next-token prediction loss.

\subsection{Pre-training}

Our goal of the pre-training stage is to endow \model{} with a broad range of robotic capabilities and strong generalization across diverse tasks, settings, modalities, and embodiments, driven by natural language instructions. To this end, we train \model{} on the Open X-Embodiment~\cite{open_x_embodiment_rt_x_2023} (OXE) dataset, comprising trajectories from different robots performing a wide range of tasks, including subsets like BridgeV2 \cite{walke2023bridgedata}, DROID \cite{khazatsky2024droid}. Similar to OpenVLA \cite{kim2024openvla}, we resized all frames to 224 x 224px for training. Details to the pre-training mixture split can be found in the Appendix.

\begin{figure}[h!]
    \centering
    \begin{subfigure}[b]{0.4\textwidth}
    \includegraphics[width=\textwidth]{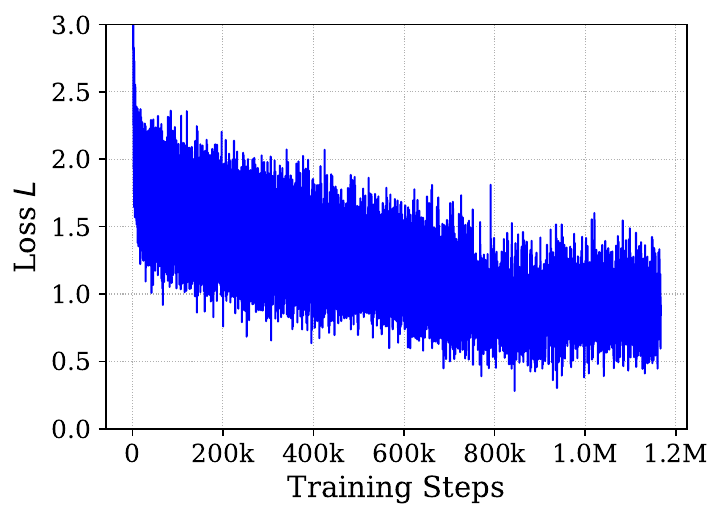}
    \caption{}
    \label{fig:train_loss}
    \end{subfigure}~
    \begin{subfigure}[b]{0.4\textwidth}
    \includegraphics[width=\textwidth]{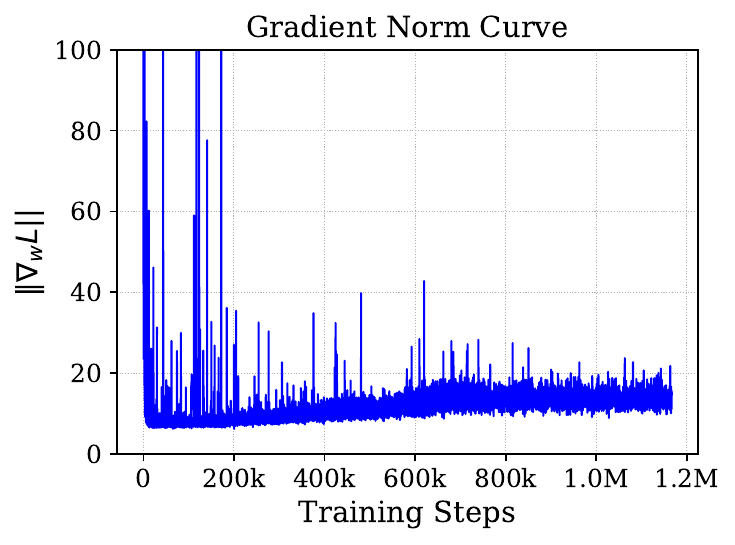}
    \caption{}
    \label{fig:grad_norm}
    \end{subfigure}
    \caption{(a) Training Loss Curve; (b) Gradient Norm Curve.}
\end{figure}

We trained \model{} for roughly three weeks on a single node of 8xH100 GPU, totaling $\sim$4000 H100 GPU hours. We used a batch size of 256 and performed 1.1 million gradient updates with the AdamW \cite{loshchilov2017decoupled} optimizer. We applied a linear warmup over the first 50k steps to a peak learning rate of $5 \times 10^{-5}$, followed by cosine decay to zero. To enhance training efficiency and reduce memory footprint, we utilized FlashAttention and trained with bf16 precision. We report the training loss and the gradient norm curve in \cref{fig:train_loss,fig:grad_norm}. The training process demonstrated a generally stable loss curve, with a downward trend with no significant spikes. While the gradient norm curve showed occasional spikes throughout training, these did not appear to disrupt the overall smooth progression of the loss.

\subsection{\modellong{}}
Several works have shown that action chunking, predicting a longer action horizon without frequent replanning leads to superior performance.\cite{zhao2023learningfinegrainedbimanualmanipulation,chi2024diffusionpolicyvisuomotorpolicy}. Motivated by these findings, we trained a variant of NORA, called \modellong{}, which uses an action chunk size of 5. \modellong{} shares the exact same architecture as NORA, but predicts an action horizon of 5 actions from a given state. We pretrained \modellong{} for 900k steps on the same pretraining dataset as \model{}.

\section{Experiments}

To investigate the efficacy of \model{} in both simulated and real-world environments as a generalist robotic control foundation model that is (i) capable of performing previously unseen tasks (zero-shot adaptation) and (ii) suitable for fine-tuning for novel robotic downstream tasks. 
\model{} is evaluated against prior state-of-the-art robotic foundation models on three different categories of tasks.

\begin{figure}[t]
    \centering
    \includegraphics[width=\textwidth]{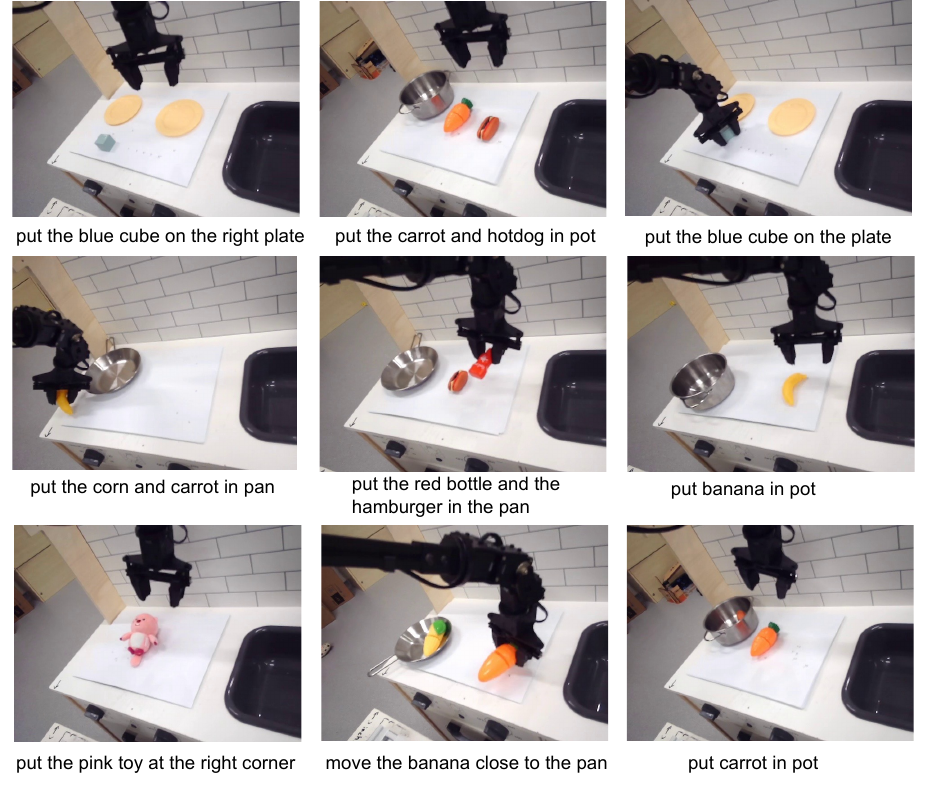}
    \caption{Real-world robot environments and task setups. We evaluate \model{} across 9 diverse tasks to assess its instruction understanding, spatial reasoning, and multi-task motion planning capabilities.}
    \label{fig:robot_setup}
\end{figure}

\subsection{Evaluation Setup and Metrics}

To evaluate the robustness of \model{} across diverse environments and robotic embodiments, we use (i) a real-world WidowX robot platform by \citet{walke2023bridgedata} and (ii) LIBERO~\cite{liu2023libero} simulation benchmark comprising 30 procedurally-generated disentangled tasks requiring a deep understanding of the varied spatial layouts (LIBERO-Spatial), objects (LIBERO-Object), and task goals (LIBERO-Goal) and 10 long-horizon entangled tasks (LIBERO-Long); this benchmark also accompany a training dataset.
In both cases, the policy model takes a third-person camera feed and a natural language instruction as input to predict the end-effector velocity actions to control the robot across 500 trials. We finetune \model{}on the corresponding dataset for 150 epochs with a batch size of 128 and a learning rate of $5 \times 10^{-5}$.

To determine the generalization capabilities of the policy model, we develop a suite of challenging evaluation tasks that involve out-of-domain (OOD) objects, spatial relationships, and multiple pick-and-place tasks, as shown in \cref{fig:robot_setup}. All the policies are assessed under identical real-world setups by ensuring consistent camera angles, lighting conditions, and backgrounds.
Each task is conducted over $10$ trials, adhering to the methodology by \citet{kim2024openvla}. 

If the robot successfully completes the task specified by the prompt, it is counted as a success (\textbf{succ}), receiving a score of $1$; otherwise, a score of $0$ is assigned:
\begin{equation*}
    \text{\% success rate} \coloneq (100 \mathtt{E}_{\tau \sim \mathcal{D}_\text{eval}} \mathtt{1}[\text{task $\tau$ is successfully completed}]) \%.
\end{equation*}

\subsection{Baselines}

For a comparative evaluation of \model, we compare its performance with the following baseline methods.

\textbf{OpenVLA} \citep{kim2024openvla}: A VLA model is built upon a Llama 2 language model\cite{touvron2023llama} combined with a visual encoder that integrates pretrained features from DINOv2 \cite{dinov2} and SigLIP\cite{zhai2023sigmoid}. It is pretrained on the Open-X-Embodiment dataset~\citep{open_x_embodiment_rt_x_2023}, which comprises 970k real-world robot demonstrations.

\textbf{SpatialVLA} \citep{spatialvla2025}: A VLA model focused on spatial understanding for robot manipulation, incorporating 3D information such as spatial movement. It learns a generalist policy for spatial manipulation across diverse robots and tasks. SpatialVLA predicts four actions at a time.

\textbf{TraceVLA} \citep{tracevla2024}: A VLA model enhancing spatial-temporal reasoning via visual trace prompting. Built by fine-tuning OpenVLA on robot manipulation trajectories, it encodes state-action history as visual prompts to improve manipulation performance in interactive tasks.

\textbf{RT-1} \citep{brohan2023rt1roboticstransformerrealworld}: A scalable Robotics Transformer model designed to transfer knowledge from large task-agnostic datasets. Trained on diverse robotic data, RT-1 achieves a high level of generalization and task-specific performance across a variety of robotic tasks, demonstrating the value of open-ended task-agnostic training of high-capacity models.

\begin{table*}
\centering
\caption{Experimental results (\% success rate) of \model{} and baselines on nine real-world WidowX-250 robot manipulation tasks.}
\resizebox{\textwidth}{!}{
\begin{tabular}{l p{7cm} c c c c c c  }
\toprule
Category&Task&RT-1&OpenVLA &SpatialVLA&\model{} (Ours)\\
\midrule
Multiple objects&Put the red bottle and the hamburger in the pan &0&20&0&\bf 40\\
Multiple objects&Put the carrot and hotdog in pot&0&0&0&\bf 30\\
Multiple objects&Put the corn and carrot in the pan&0&\bf 30&0&\bf 30\\
\midrule
OOD object&put carrot in pot&0&80&20 &\bf 90\\
OOD object&Put banana in pot&1&40&0&\bf 90\\
OOD object&Put the blue cube on the plate&0&50&0&\bf 70\\
\midrule
Spatial&Put the pink toy at the right corner&0&\bf 60&30&\bf 60\\
Spatial&Put the blue cube on the right plate&0&\bf 30&0&20\\
Spatial&Move the banana close to the pan&30&50&50&\bf 80\\

\midrule
\multicolumn{2}{l}{\hspace{2.5cm}\textbf{Average}} &4.4&40&11.1&\bf 56.7\\
\bottomrule
\end{tabular}
}
\label{tab:real_robot_results}
\end{table*}

\begin{figure}[t!]
        \centering
         \vspace{-0.8em}
        \begin{tikzpicture}
    \definecolor{customRed}{HTML}{F5867F}
    \definecolor{customYellow}{HTML}{FFBC80}
    \definecolor{customBlue}{HTML}{6B98C4}
    \definecolor{customPurple}{HTML}{EEBEC0}
     \definecolor{customgreen}{HTML}{A8B9A4}
        \begin{axis}[
            ybar,
            bar width=.4cm,
            width=0.6\textwidth,
            height=4.5cm,
            enlarge x limits=0.15,
            ylabel={Avg Success Rate ($\mathbf{\%}$) },
            xlabel={Different Categories of Tasks},
            symbolic x coords={ Multiple, OOD Object, Spatial},
            xtick=data,
            ymin=0,
            ymax=100,
            ytick={0,10,20,30,40,50,60,70,80,90,100},
            grid=major,
            xmajorgrids=false, 
            tick label style={
        font=\fontsize{6}{1}\selectfont 
    },
    xlabel style={
        font=\fontsize{9}{1}\selectfont 
    },
    ylabel style={
        font=\fontsize{9}{1}\selectfont 
    },
            grid style={dashed,gray!30},
                    legend style={
                font=\fontsize{7}{1}\selectfont, 
                legend style={row sep=-0.1cm},
                at={(1,1)},
            },
            legend image code/.code={
              \draw[#1] (0cm,-0.1cm) rectangle (0.15cm,0.1cm);
            }, 
        ]
        \addplot [fill=customgreen] coordinates {(Multiple,0) (OOD Object,3.3)  (Spatial, 10)  };
        \addplot [fill=customRed] coordinates {(Multiple,0) (OOD Object,6.7)  (Spatial, 26.7)  };
        \addplot [fill=customYellow] coordinates {(Multiple,16.7) (OOD Object,56.7) (Spatial,46.7)  };
        \addplot [fill=customBlue] coordinates {(Multiple,33.3) (OOD Object,83.3)  (Spatial,53.3)};
        \legend{RT-1,Spatial VLA,OpenVLA,  \model}   
        \end{axis}
    \end{tikzpicture}
    \vspace{-1.2em}
    \caption{Experimental results on different categories of real-world robot tasks.}
    \vspace{-0.8em}
    \label{fig:real_robot_result_overall}
\end{figure}
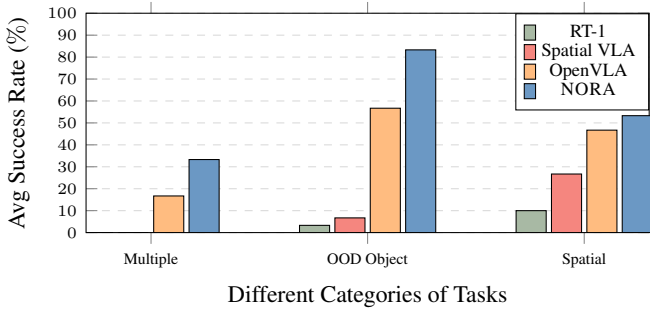

\subsection{Experimental Outcomes}

\paragraph{Improved policy generation under real-world settings.}
The experimental results in \cref{fig:real_robot_result_overall} demonstrate the significant superiority of \model{} in policy generation over the baselines across three types of tasks: out-of-domain object grasping, tasks requiring spatial reasoning, and multi-object grasping. Specifically, as shown in \cref{tab:real_robot_results}, \model{} achieves impressive success rates in OOD/zero-shot object grasping tasks, such as ``\textit{put the carrot in pot}'' and ``\textit{put banana in pot}'' with a success rate of up to 90\%. This significantly outperforms the baseline models that struggle at these tasks.
Similarly, at tasks that require spatial reasoning, such as, ``\textit{put the pink toy at the right corner}'' and ``\textit{move the banana close to the pan}'', \model{} generally shows superior performance. While SpatialVLA includes modules designed to capture 3D spatial features, we found that, despite its ability to correctly determine spatial orientation, its performance in object grasping is worse. This limitation often results in task failures, as the model struggles to complete the necessary manipulation, even when spatial relationships are understood correctly.

\model{} outperforms the baselines at multi-object grasping tasks, but the performance advantage is much more narrow as compared to the prior two task types. At tasks like ``\textit{Put the red bottle and the hamburger in the pan}'' and ``\textit{Put the carrot and hotdog in pot}'', the success of \model{} appears to be much more precarious at below $50\%$, indicating \ul{a substantial room for improvement in the tasks requiring handling of multiple objects}.

Overall, our approach exhibits robust performance across a variety of task settings, showcasing superior generalization capabilities compared to the baseline models.

\definecolor{Gray}{gray}{0.95}
\begin{table*}
\centering
\caption{Experimental results (\% success rate) of \model{} and baselines on LIBERO Simulation Benchmark. Each method is evaluated on four task suites over 500 trials. Fine-tuned \model{}-Long achieves the best overall performance. Results marked with $\ast$ are from SpatialVLA \cite{spatialvla2025}. \texttt{AC} indicates the use of action chunking strategy.} 
\resizebox{\textwidth}{!}{
\begin{tabular}{l c c c c >{\columncolor{Gray}}c}
\toprule
Models&LIBERO-Spatial&LIBERO-Object &LIBERO-Goal&LIBERO-Long&Average\\
\midrule
\texttt{OpenVLA fine-tuned} $\ast$ &84.7&88.4&79.2&53.7&76.5\\
\texttt{TraceVLA fine-tuned} $\ast$ &84.6&85.2&75.1&54.1 &74.8\\
\texttt{\model{}-fine-tuned} (Ours) &85.6&87.8&77&45&73.9\\

\midrule
\texttt{SpatialVLA fine-tuned-AC} $\ast$ &88.2&89.9&78.6&55.5&78.1\\
\texttt{\model{}-fine-tuned-AC} (Ours) &85.6&89.4&80&63&79.5\\
\texttt{\model{}-Long-fine-tuned} (Ours) &\textbf{92.2}&\textbf{95.4}&\textbf{89.4}&\textbf{74.6}&\textbf{87.9}\\

\bottomrule
\end{tabular}
}
\label{tab:libero_sim}
\end{table*}

\paragraph{Improved performance in simulated environment.}

To economically evaluate the adaptability of \model{} to new robot embodiments, we employ LIBERO Simulation Benchmark~\citep{liu2023libero}. Firstly, \texttt{NORA-fine-tuned} are obtained by fine-tuning the pretrained \model{} on the LIBERO training dataset. The fine-tuning objective of \texttt{\modellong{}} is long-horizon planning, predicting the next five actions at each step, instead of only the next action for \texttt{NORA-fine-tuned}.

As shown in \cref{tab:libero_sim}, \texttt{\modellong{}} achieves the highest average success rate (87.9\%) across all methods, demonstrating strong generalization in both short- and long-horizon scenarios. Among the fine-tuned baselines without action chunking , OpenVLA achieves the best average (76.5\%). \model{}, demonstrates comparable performance to OpenVLA in spatial, object, and goal-related tasks, but it falls short in long-horizon scenarios.

Notably, when both NORA variants are fine-tuned with action chunking, there is a significant increase in the LIBERO-Long success rate, emphasizing the importance of action chunking for long-horizon tasks. \texttt{\modellong{}} especially excels on LIBERO-Long, achieving a success rate of 74.6\%, showcasing its ability to reason over extended temporal windows. These results highlight the effectiveness of our model in adapting to new environments and reinforce the utility of windowed training for long-horizon policy generalization.".

\begin{figure}[t]
    \centering
    \includegraphics[width=0.95\textwidth]{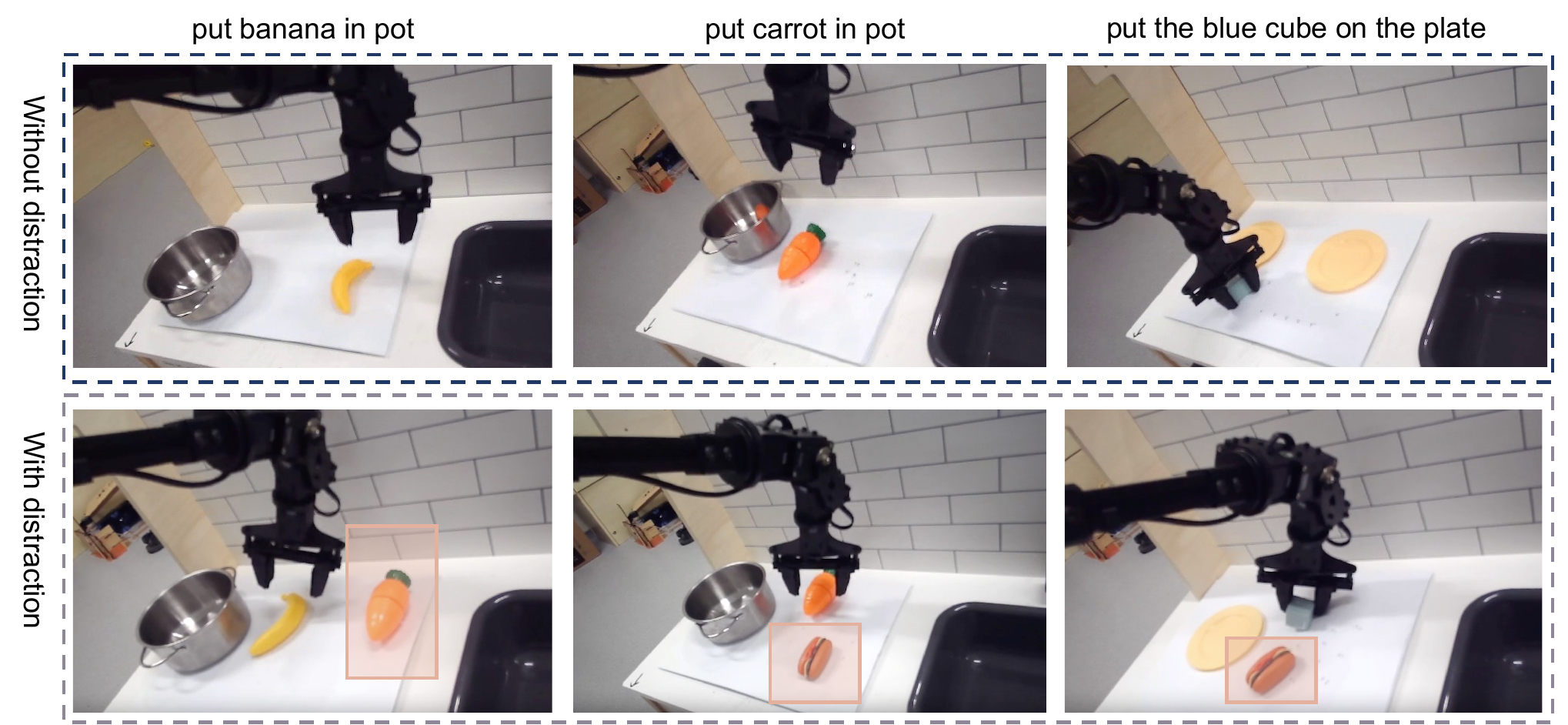}
    \caption{Comparison of tasks with and without distraction.}
    \label{fig:distraction_object}
\end{figure}

\begin{figure}[t!]
    \centering
    \vspace{-0.8em}
    \begin{tikzpicture}
        \definecolor{customBlue}{HTML}{6B98C4}
        \definecolor{customYellow}{HTML}{FFBC80}
        \begin{axis}[
            ybar,
            bar width=.55cm,
            width=0.5\textwidth,
            height=4.5cm,
            enlarge x limits=0.3,
            ylabel={Avg Success Rate ($\mathbf{\%}$)},
            symbolic x coords={OpenVLA, NORA},
            xtick=data,
            ymin=0,
            ymax=100,
            ytick={0,10,20,30,40,50,60,70,80,90,100},
            grid=major,
            bar shift=-0.15cm, 
            legend style={
                font=\fontsize{7}{1}\selectfont,
                at={(0.5,1.15)},
                anchor=south,
                legend columns=2
            },
            tick label style={
                font=\fontsize{7}{1}\selectfont
            },
            xlabel style={
                font=\fontsize{9}{1}\selectfont
            },
            ylabel style={
                font=\fontsize{9}{1}\selectfont
            },
            grid style={dashed,gray!30},
            legend image code/.code={
              \draw[#1] (0cm,-0.1cm) rectangle (0.15cm,0.1cm);
            },
        ]
        \addplot+[fill=customBlue, bar shift=-0.55cm] coordinates {(OpenVLA,56.67) (NORA,83.33)};
        \addplot+[fill=customYellow, bar shift=0.15cm] coordinates {(OpenVLA,50) (NORA,56.67)};
        \legend{w/o Distractors, w/ Distractors}
        \end{axis}
    \end{tikzpicture}
    \vspace{-1.2em}
    \caption{Success rates of OpenVLA and NORA with and without distractors.}
    \vspace{-0.8em}
    \label{fig:distractors_comparison}
\end{figure}
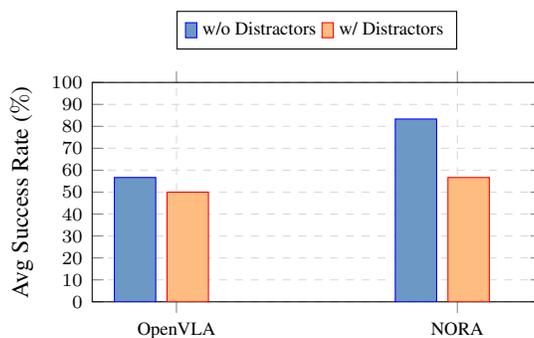

\paragraph{Distractions in the environment.}
To better simulate real-world environments, we selected three straightforward tasks (\cref{fig:distraction_object}) where both OpenVLA and \model{} initially perform well, as shown in \cref{tab:real_robot_results}. We then introduced additional objects into the environment to serve as distractions. As illustrated in \cref{fig:distractors_comparison}, both policies experienced significant performance drop in the presence of these distractions, highlighting their fragility.

\paragraph{Action chunking performs worse on WidowX.} 
To test if action chunking is effective in our robotic embodiment, we evaluated \modellong{} by selecting one task from each of the three categories: (\textit{put the carrot in the pot}''), (\textit{put the red bottle and hamburger in the pot}''), and (``\textit{Put the pink toy at the right corner}''). We first experimented by executing all 5 predicted actions sequentially without replanning. However, we observed that the Widow X robot often crashed into the environment, as the accumulated actions tended to result in excessively large movements. Similarly, SpatialVLA also exhibits similar behavior of crashing into the environment when executing all actions predicted (4 actions) at the same time.

Next, we evaluated \modellong{} by executing only the first action from each predicted action chunk. This approach resolves the issue of the robot crashing into the environment and achieves a success rate of 80$\%$ on the (`\textit{put carrot in the pot}) task. However, when evaluated on multi-object pick-and-place tasks, \modellong{} always stops moving after successfully placing the first object into the pot, resulting in a final success rate of 0$\%$ for multi-object pick-and-place. Lastly, we evaluate on the spatial category task. We observe \modellong{} achieve a 70$\%$ success rate (``\textit{Put the pink toy at the right corner}''), demonstrating similar performance to \model{}.

Interestingly, when comparing \model{} to \modellong{}, we observe a few key differences. Notably, \modellong{} estimates affordance points differently, consistently attempting to grip objects from the side — specifically around the 2 o'clock direction - whereas \model{} tends to grip objects directly from above. While gripping objects from the sid does not significantly impact the grasping of larger objects, it makes smaller objects much harder to pick up. We further evaluated performance on another spatial task (``\textit{Move the banana close to the pan}'') and found that \modellong{} struggled to grasp the banana due to poor affordance point estimation. This made it difficult to pick up the smaller banana object, resulting in a final success rate of 40$\%$.

Hence, we find that \modellong{} is less robust than \model{}, as it struggles to complete multi-object pick-and-place tasks and exhibits poorer affordance point estimation, leading to difficulties in grasping smaller objects.

\paragraph{Action chunking improves performance in simulation.}
We hypothesize that action chunking is more effective when operating at higher control frequencies. For example, Diffusion Policy \cite{chi2024diffusionpolicyvisuomotorpolicy} predicts robot commands at 10 Hz, but these commands are interpolated to 125 Hz for execution. Similarly, OpenVLA-OFT+ \cite{kim2025finetuningvisionlanguageactionmodelsoptimizing} also employs action chunking and demonstrates improvements in real-world ALOHA tasks \cite{zhao2023learningfinegrainedbimanualmanipulation}, which operates at 25 Hz. Since we currently lack access to a robotic embodiment capable of high-frequency control, we evaluate this hypothesis in the LIBERO simulation environment, which operates at 20 Hz.

We finetuned \model{} and \modellong{} model on the LIBERO simulation benchmark with both action chunk size of 5, obtaining \texttt{\model{}-fine-tuned-AC} and \texttt{\model{}-Long-fine-tuned}. We observe that \texttt{\model{}-fine-tuned-AC} achieves substantially higher performance across all aspects of the LIBERO simulation compared to \texttt{\model{}-fine-tuned}, along with a higher average success rate than other variants. Notably, \texttt{\model{}-Long-fine-tuned} significantly outperforms all baselines, highlighting the effectiveness of pretraining with action chunking and transferring the long horizon planning ability to the downstream task. However, a key to note is that LIBERO is a simulation environment that does not necessarily represent \model{}-Long will achieve superior performance compared to \model{} in high-frequency real world robotic tasks.

\subsection{Case Study}
We conduct real-world comparisons between \model{}, OpenVLA, and SpatialVLA across three task categories: (1) out-of-distribution (OOD) object manipulation (``\textit{put carrot in the pot}''), (2) spatial relationship reasoning (``\textit{move the banana close to the pan}''), and (3) multi-object pick-and-place (``\textit{put the red bottle and the hamburger in the pan}'').

In the first case, which involves an OOD object, both \model{} and OpenVLA successfully complete the task. However, SpatialVLA fails due to incorrect affordance point estimation, resulting in an unsuccessful grasp.

In the second case, requiring spatial reasoning, OpenVLA fails to follow the instructions correctly despite grasping the object, showing limitations in directional understanding. \model{} successfully places the banana near the pan as instructed, while SpatialVLA exhibits unstable performance due to poor grasp strategy.

In the third case, which requires handling multiple objects, baseline models fail to execute the task reliably. For instance, SpatialVLA attempts to grasp suboptimal locations or orientations, leading to failed pickups. In contrast, \model{} completes the task successfully by accurately identifying and manipulating both objects.

These case studies highlight the robustness of \model{} across diverse and challenging real-world scenarios. It demonstrates reliable performance in novel object manipulation, spatial reasoning, and multi-object tasks, where baseline models often struggle due to affordance errors, inadequate spatial understanding, or unstable grasp execution.

\begin{figure}[t]
    \centering
    \includegraphics[width=1\textwidth]{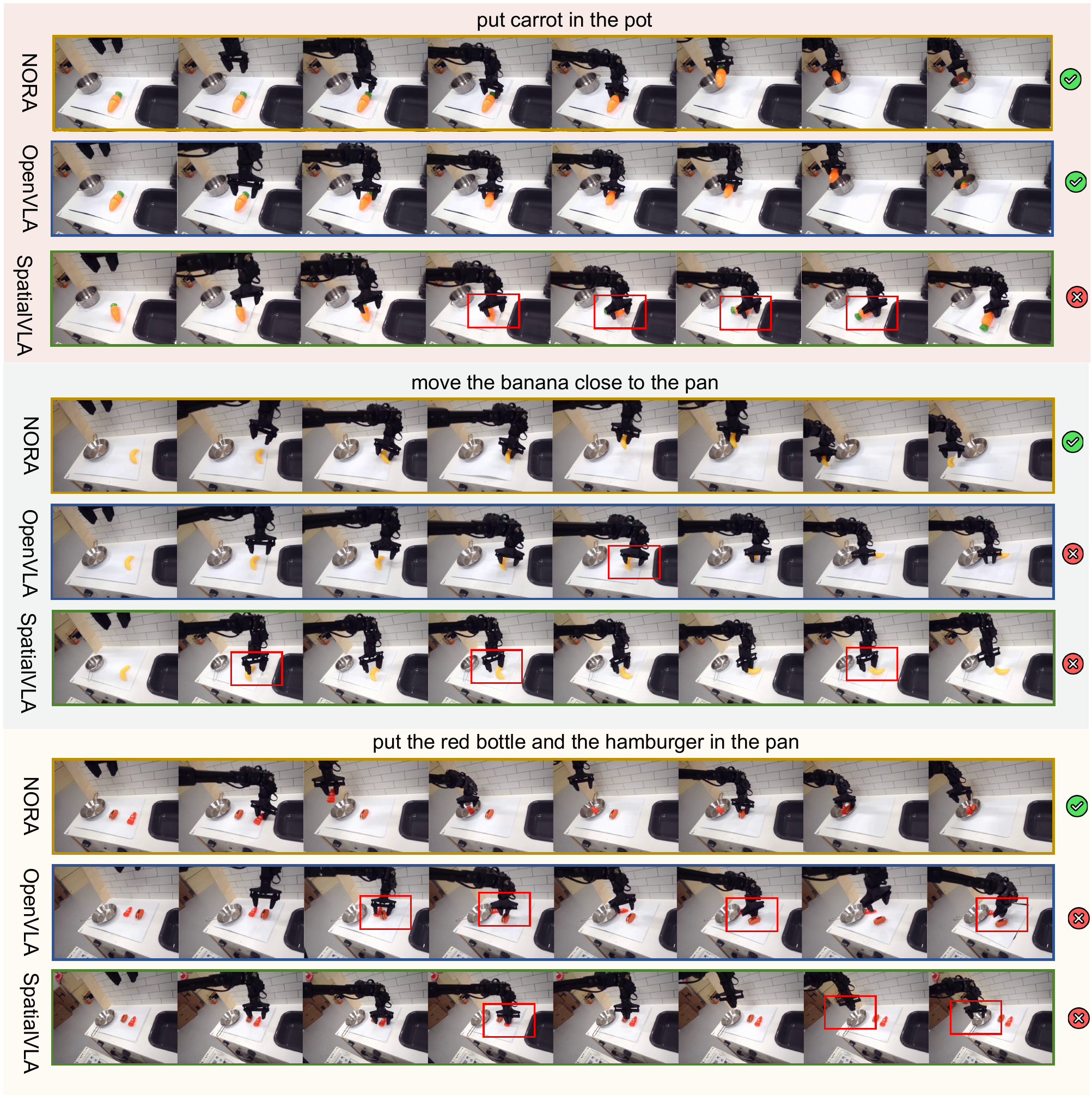}
    \caption{Case study comparisons of \model{} and baseline methods in real-world robotic tasks.}
    \label{fig:case_study}
\end{figure}

\section{Related Works}
\label{sec:rw}

\paragraph{Generalist Robot Policies.}
Robotic learning has increasingly advanced towards training generalist policies capable of executing diverse tasks across multiple embodiments \cite{brohan2023rt1roboticstransformerrealworld, brohan2023rt2visionlanguageactionmodelstransfer, ebert2021bridgedataboostinggeneralization, walke2023bridgedata, open_x_embodiment_rt_x_2023, octo_2023}.
Octo \cite{octo_2023} adopts a compositional learning framework to support multi-task control, while RT-1 \cite{brohan2023rt1roboticstransformerrealworld} demonstrates how large-scale robot demonstrations can be used to train scalable behavior policies across tasks and embodiments.
These systems show the importance of combining data diversity and modular architectures to support robust general-purpose robot control.

\paragraph{Vision-Language-Action Models.}
Vision-Language-Action (VLA) models extend vision-language models by integrating robot actions into the token space, allowing them to directly generate low-level controls from multimodal inputs \cite{brohan2023rt2visionlanguageactionmodelstransfer, open_x_embodiment_rt_x_2023, kim2024openvla, driess2023palmeembodiedmultimodallanguage}.
RT-2 \cite{brohan2023rt2visionlanguageactionmodelstransfer} combines Internet-scale vision-language data with robot trajectory datasets to enable scalable visuomotor learning.
RT-2-X \cite{open_x_embodiment_rt_x_2023} further extends this paradigm by scaling the model to 55B parameters and training it on the large and diverse Open X-Embodiment dataset.
OpenVLA \cite{kim2024openvla} improves model accessibility and reproducibility by training an open-source 7B-parameter VLA on over 970k real-world robot demonstrations, using a modular combination of pretrained visual encoders and language models.

To enhance spatial reasoning capabilities, SpatialVLA \cite{spatialvla2025} introduces Ego3D position encoding and adaptive action grids to encode 3D spatial information, allowing generalist policies to better reason about object locations and affordances across different robots and tasks.
TraceVLA \cite{tracevla2024} focuses on temporal grounding, encoding historical state-action pairs as visual prompts to improve policy effectiveness in interactive manipulation sequences.

Beyond spatial and temporal representations, recent works have explored the incorporation of intermediate reasoning steps to address complex planning problems.
CoT-VLA \cite{cotvla2025} introduces visual chain-of-thought reasoning by autoregressively generating future visual goals before producing action sequences to achieve them.
This explicit decomposition into intermediate visual states enables the model to exhibit enhanced temporal planning and interpretability.

In parallel, $\pi_0$ \cite{pi02024} proposes a flow-matching framework atop a pretrained vision-language model, enabling it to inherit semantic grounding while supporting general robot control across single-arm, dual-arm, and mobile platforms.
The model is trained on a large and diverse robot dataset and evaluated on a wide range of manipulation tasks, including dexterous operations like laundry folding and box assembly.

Despite recent advancements, current VLAs often underutilize one of the most valuable capabilities of their underlying language and vision-language model's ability to reason through the sequential steps required to solve complex tasks.

\section{Conclusion}
We introduce NORA, a 3B-parameter Visual-Language-Action (VLA) model designed to optimize robotic task execution by reducing computational overhead and improving efficiency. NORA is trained on the Open X-Embodiment dataset, which incorporates diverse robotic embodiments and tasks. We adopt Qwen-2.5-VL-3B as the backbone of NORA, due to its outstanding performance in vision-language understanding, which provides significant gains in multimodal reasoning and task execution. To enhance the robot's prediction speed and improve action encoding/decoding efficiency, we apply the FAST+ tokenizer to discretize continuous action tokens. The experimental results demonstrate that NORA outperforms existing VLA models, showing significant improvements in task performance, especially in real-world environments.

\bibliography{custom}
\bibliographystyle{iclr2025_conference}

\appendix
\section{Appendix}
You may include other additional sections here.

\end{document}